\documentclass[sigconf]{acmart}

\renewcommand\footnotetextcopyrightpermission[1]{}
\AtBeginDocument{%
  }

\setcopyright{acmlicensed}
\copyrightyear{2025}
\acmYear{2025}
\acmDOI{XXXXXXX.XXXXXXX}
\acmConference[SIGIR '25]{The 48th International ACM SIGIR Conference on Research and Development in Information Retrieval}{July 13--17, 2025}{Padova, Italy}
\usepackage{graphicx}
\usepackage{float}
\usepackage{mdframed}
\usepackage{fancyvrb} 
\usepackage{caption} 
\usepackage{graphicx}
\usepackage{booktabs}
\usepackage{caption}
\usepackage{subcaption}
\usepackage{tabularx} 
\usepackage{booktabs} 
\usepackage{booktabs}
\usepackage{multirow}
\usepackage{array}
\usepackage{tcolorbox}
\tcbuselibrary{skins,breakable} 

\usepackage{tcolorbox}
\newtcolorbox{promptbox}{
    colback=gray!10,    
    colframe=gray!50,   
    boxrule=0.5pt,      
    arc=3pt,            
    width=\columnwidth, 
    left=6pt,           
    right=6pt,
    top=4pt,
    bottom=4pt,
    boxsep=0pt,
    fontupper=\small
}

\newtcolorbox{widepromptbox}[1][]{
    colback=gray!10,       
    colframe=gray!50,      
    boxrule=0.5pt,         
    arc=3pt,               
    width=\textwidth,      
    left=12pt,             
    right=12pt,
    top=8pt,
    bottom=8pt,
    boxsep=0pt,
    enhanced,              
    before skip=12pt,      
    after skip=12pt,       
    fontupper=\small,
    #1                      
}
\begin{document}

\title{Knowledge-Aware Diverse Reranking for Cross-Source Question Answering}


\author{Tong Zhou}
\affiliation{%
  \institution{Institute of Automation, Chinese Academy of Sciences}
  \city{Beijing}
  \country{China}
}
\email{tong.zhou@ia.ac.cn}


\begin{abstract}
This paper presents Team Marikarp's solution for the SIGIR 2025 LiveRAG competition. The competition's evaluation set, automatically generated by DataMorgana from internet corpora, encompassed a wide range of target topics, question types, question formulations, audience types, and knowledge organization methods. It offered a fair evaluation of retrieving question-relevant supporting documents from a 15M documents subset of the FineWeb corpus. Our proposed knowledge-aware diverse reranking RAG pipeline achieved first place in the competition.

\end{abstract}

\maketitle

\section{Introduction}
The SIGIR 2025 LiveRAG Challenge represents a systematic initiative for evaluating modern Retrieval-Augmented Generation \citep{gao2023retrieval, fan2024survey} systems in open-domain question answering scenarios. Organized by the Technology Innovation Institute, this competition provides a standardized benchmark comprising two core components: (1) a fixed document corpus combining FineWeb datasets \citep{penedo2024the} containing 15M documents of cleaned web data, (2) the Falcon3-10B-Instruct foundation model \citep{Falcon3} as the mandatory LLM component.

Participants are tasked with developing RAG systems that integrate retrieval from either custom-built indices or pre-configured options: a Pinecone dense index utilizing E5-base \citep{wang2022text} embeddings on 512-token chunks, and an OpenSearch sparse index implementing BM25. A notable feature of the competition framework is DataMorgana \citep{filice2025generating}, a synthetic question generator that produces adversarial test queries through controlled perturbations of original document content.

\subsection{Evaluation}
The evaluation dataset for LiveRAG is a 500-item test set generated by DataMorgana, a lightweight two-stage tool designed for creating diverse synthetic Q\&A benchmarks. Leveraging natural language descriptions, DataMorgana allows detailed configurations of user categories (e.g., expert, novice, domain-specific roles like patients or researchers) and question categories (e.g., factoid, open-ended, with/without premise, varying phrasing styles such as concise-natural or long-search-query), enabling combinatorial diversity across lexical, syntactic, and semantic dimensions. The generation process involves sampling documents from the FineWeb corpora, then invoking an Claude-3.5-Sonnet with prompts tailored to the configured categories to produce questions that reflect realistic user interactions. 
The test questions include single-document type and multi-document type. The multi-document type questions are generated by DataMorgana based on two relevant documents, and answering such questions must simultaneously reference both documents.
This dataset ensures high coverage of diverse query types, from fact-based inquiries to complex, premise-including questions, and incorporates filtering to validate faithfulness to source documents, providing a robust and customizable benchmark for evaluating RAG systems in the LiveRAG challenge.

Evaluation employs a multi-stage verification protocol: \textbf{Relevance Scoring}: Answers are rated on a 4-level scale assessing both correctness and concision (scores: -1, 0, 1, 2). \textbf{Faithfulness Verification}: Responses are checked against retrieved evidence using a 3-tier grading system (scores: -1, 0, 1). The assessment pipeline combines automated evaluation using Claude-3.5 Sonnet for initial scoring with human verification of top submissions. 


To address bias and misinformation in RAG reranking, we propose a knowledge-aware diverse reranking RAG pipeline. This paper details its architecture, training, and experimental results.

\section{Pipeline}
This section outlines our Retrieval-Augmented Generation (RAG) pipeline, focusing on its three main stages: retrieval, reranking, and answer generation.

\subsection{Retrieval}
Our retrieval strategy combines sparse and dense methods to maximize recall. 
Questions generated by DM can vary significantly, ranging from direct inquiries about the target document to questions requiring substantial rephrasing of keywords or entities. 
To address this, we employ a hybrid approach. We treat the question directly as a query and retrieve the top $N_{ret}$ results from both sparse $M_{ret}^{sparse}$ and dense retrievers $M_{ret}^{dense}$. 
The top $N_{ret}$ results from each are then merged by alternately selecting documents from their respective ranked lists, preserving intrinsic orderings, and discarding duplicates until a final list of $N_{ret}$ unique documents is obtained.

\subsection{Reranking}
Given the inherent limitations of LLM regarding context window length and their susceptibility to noise content, the number of documents $N_{ans}$ provided for answer generation must be significantly smaller than the initially retrieved set $N_{ret}$. Consequently, an effective reranking mechanism is critical for prioritizing the most relevant documents and ensuring optimal LLM performance.

\subsubsection{Initial Reranking}

We initially employ a large-scale pre-trained reranking model, denoted as $M_{rank}$, to perform a preliminary reranking. This model processes a query $Q$ and document $D$ pair and outputs a relevance score $S_{rank}$ between 0 and 1. $S_{rank}=M_{rank}(Q, D)$. We rank all retrieved documents according to these scores.

\subsubsection{Knowledge Element Declaration}


However, relying solely on the complete query for reranking can introduce a bias towards specific knowledge facets, particularly within multi-document contexts. For instance, consider the query: \textit{``Who had more world titles, Muhammad Ali or Ramon Dekkers?''} If the semantic encoding of the query overemphasizes one athlete's knowledge or if more relevant information is retrieved for one athlete concerning \textit{world titles}, the reranking might disproportionately favor that individual. This can result in a ranked list that skews towards one aspect of the knowledge, failing to provide all necessary information to answer the complete question.
Therefore, to mitigate this potential bias and ensure a more balanced information synthesis, we propose leveraging the semantic understanding capabilities of a Large Language Model (LLM) to decompose the initial query into independent sub-queries. This decomposition facilitates a more granular and targeted reranking process.
Nevertheless, the efficacy of such LLM-based decomposition hinges on a nuanced understanding of the query's underlying knowledge requirements. Without this, an LLM might misinterpret the query's intent, leading to suboptimal sub-query generation. For example, when presented with the question, \textit{``What is the difference between llama and falcon in training methods?''}, an LLM lacking relevant contextual information might erroneously interpret \textit{llama} and \textit{falcon} as animal species, decomposing the query into \textit{``How to train a llama?''} and \textit{``How to tame a falcon?''}. Conversely, by integrating effective joint semantic retrieval, the LLM can ascertain from the retrieved corpus that \textit{llama} and \textit{falcon} refer to large language models. This informed understanding enables a more accurate decomposition of the original query, aligning the sub-queries with the user's true informational needs.

To address the \textbf{bias issue} caused by reranking based on the original question and the \textbf{misleading problem} due to large language models' lack of understanding of the question context, we propose a knowledge-aware diverse reranking strategy.
We utilize the LLM $M_{know}$ referencing the top documents (top $N_{know}$ documents) from the preliminary reranking one by one to deconstruct the question in order to identify essential knowledge elements ${E_{q}}_{i}$ required for a comprehensive answer. 
Simultaneously, it extracts knowledge elements ${E_{d}}_i$ already present within each document $D_i$ to guarantee the understanding of the given document, thereby assessing the document's specific content coverage.
These can also used for our heuristic automatic judgment of extraction quality.
Specifically, the complete knowledge set ${E_{q}}_{i}$ must necessarily include all information extracted from the current document $D_i$ that is critical for addressing the target question. This implies ${E_{d}}_i \subseteq {E_{q}}_{i}$. Figures \ref{fig:prompt_dec} and \ref{fig:demo_dec} present an example prompt and a corresponding demonstration, respectively.

\subsubsection{Knowledge Element Summarization}

Recognizing that individual documents provide partial or aspect-specific information, our strategy further leverages the LLM $M_{summ}$ to synthesize the declarative knowledge ${E_{q}}$ aggregated from these top documents. This synthesis culminates in two distinct, yet complementary, knowledge representations, ${E_{q}}^{sum_0}$ and ${E_{q}}^{sum_1}$, designed to be core to the query's intent while offering diverse and comprehensive coverage of the relevant information landscape. Figure \ref{fig:prompt_summ} displays the prompt used, while Figure \ref{fig:demo_summ} provides a representative example.

\subsubsection{Knowledge Aware Diverse Reranking}

We then rerank the top $N_{rank}$ documents based on these two summarized knowledge elements ${E_{q}}^{sum_0}$ and ${E_{q}}^{sum_1}$ as queries. The reranking model $M_{rank}$ is again employed, taking question concat with knowledge element as query, to produce two reranked lists focused on different aspects. 
$S_{rank}^{E_i}=M_{rank}(Q ; {E_{q}}^{sum_i}, D)$.
Same as merging two retrieved lists from sparse and dense retrievers, the top-ranked results from these two lists $S_{rank}^{E_i}$ are then interleaved by alternately selecting entries while preserving their inherent order and eliminating duplicates, ultimately producing a consolidated list of $N_{rank}$ unique documents. 

\subsection{Answer Generation}

To mitigate the impact of noise in the retrieved context, we select the top $N_{ans}$ reranked documents and provide them to the LLM $M_{ans}$. We explicitly prompt the LLM to generate a concise answer, as detailed in Figure \ref{fig:prompt_answer}.

\begin{figure}
\begin{promptbox}
\textbf{System:}\\
You are a helpful assistant.\\
You are given a question and a list of documents.\\
You need to answer the question based on the documents.\\
Please answer the question concisely, with no more than 200 words.\\

\textbf{User:}\\
Question: \{question\}\\
Context: \{documents\}
\end{promptbox}
\caption{The prompt for answer generation.}
\label{fig:prompt_answer}

\end{figure}

\section{Training and Implementation}

\subsection{DataMorgana Data Generation}
While existing DataMorgana provides examples for multiple question types, we enhance RAG system robustness through systematic problem-type expansion.
Specifically, we leverage DeepSeek-R1 \citep{guo2025deepseek}, a state-of-the-art deep reasoning LLM, to brainstorm and augment the existing single-document and multi-document problem types. 
The prompt is shown in Figure \ref{fig:prompt_brain}. 
We employ a multi-generation process followed by manual selection to curate high-quality expansions. This process ultimately expanded the number of single-document question categories to 21 and multi-document problem types to 7. 
It is worth noting that single-document question types can still be applied to multi-document question generation. For example, when generating multi-document questions, the generated questions can be required to be \textit{open-ended} and specified as \textit{comparative} questions of the multi-document type.
Further details on the expanded problem types are provided in Table \ref{tab:question_single} and \ref{tab:question_multi}. For evaluation, we generated 300 test questions for both single- and multi-document scenarios, and 3500 training questions for each.

\subsection{Knowledge Declaration Training}

To refine the model's capability in analyzing knowledge elements, we employ a supervised fine-tuning (SFT) approach leveraging rejection sampling \citep{tong2024dart, khaki2024rs}. To mitigate the alignment gap typically observed with external stronger teacher models and to reduce operational costs, we adopt a self-generation strategy. 
Specifically, the Falcon-10B-Instruct model, which is the subject of our fine-tuning, is utilized to generate the training data. 
This process involves tasking the model with processing inputs from our training set and subsequently applying heuristic rules to sample high-quality outputs.

First, we establish a typology for document relevance based on documents retrieved for questions within the training set, defining three categories: Fully Supporting, Partially Relevant, and Irrelevant.

\textbf{Fully Supporting}: A document segment (chunk) is labeled Fully Supporting if, for single-document questions, it is the first chunk retrieved via sparse retrieval (based on the concatenation of question and answer pairs as a query) that corresponds to the gold document. For multi-document questions, the first correctly retrieved chunk from each of the two gold documents (identified via sparse retrieval using the concatenation of question and answer pairs) is concatenated to form the Fully Supporting evidence. Preliminary experiments demonstrate that when utilizing a concatenated question-answer pair as the query, 97\% of multi-document questions successfully retrieve the gold chunks within the top 400 retrieved passages. Furthermore, for single-document questions, 100\% retrieval of the gold chunk is achieved within the top 100 passages.

\textbf{Partially Relevant:} A chunk is designated Partially Relevant exclusively for multi-document questions, where any single correctly retrieved chunk from a gold document falls into this category.

\textbf{Irrelevant:} Chunks are those not originating from the gold document(s). It is important to note that while these documents differ from the gold standard, they may still contain knowledge pertinent to the question.

Building upon these document labels, we implement a series of rejection rules to filter the disqualified outputs generated by Falcon-10B-Instruct. If an output fails this review, the model is re-prompted to generate, up to a maximum of retry attempts $N_{rs}$. The acceptance criteria include:
Formatting Compliance: The output must adhere to predefined structural requirements, including a discernible thought process and the final answer in JSON format. Furthermore, the output must be parsable by regular expressions to extract a list of strings.
Knowledge Quantity and Uniqueness: The number of distinct knowledge pieces identified in the output must meet specified thresholds and each piece must be unique.
Knowledge Attribution: The knowledge extracted and attributed to a given document must originate entirely from the information required to answer the current question, verified by exact string matching.
Relevance-contingent Knowledge Coverage:
For Fully Supporting documents, the extracted knowledge must be identical to the required knowledge.
For Partially Relevant documents, the extracted knowledge must be a non-empty, strict subset of the required knowledge.
For Irrelevant documents, the extracted knowledge must be a strict subset of the required knowledge and may be empty.
This iterative generation and filtering process yielded a substantial corpus of data, from which we selected 1,000 Fully Supporting, 2,500 Partially Relevant, and 6,500 Irrelevant instances that satisfied all criteria, forming our SFT dataset with 10k samples. The Falcon model was subsequently fine-tuned for 3 epochs using Low-Rank Adaptation (LoRA) \citep{hu2022lora}. LoRA was applied to all model parameters, configured with rank $r=16$, a learning rate of $1e-4$, and a batch size of $8$. After the SFT process, we get $M_{know}$.

\subsection{Parameter Selection and Experiments}

\begin{table}[ht]
\centering
\begin{tabular}{l@{} c@{} c@{} c@{} c@{} c@{} c@{} c@{}}
\toprule
Retriever      & R@10   & \ R@100  & \ R@200  & \ R@400  & \ R@1k & \ R@2k & \ R@4k \\
\midrule
sparse\&dense  & 0.41   & 0.69   & 0.73   & 0.79   & 0.86   & 0.90   & 0.93   \\
dense          & 0.37   & 0.64   & 0.69   & 0.76   & 0.83   & 0.87   & 0.91   \\
sparse         & 0.36   & 0.59   & 0.65   & 0.72   & 0.78   & 0.82   & 0.86   \\
\bottomrule
\end{tabular}
\caption{Recall of gold documents under multi-document dataset using different retrieval methods. If only one gold document is recalled, the recall rate is 50\%.}
\label{tab:recall_multi}
\end{table}

\begin{table}[ht]
\centering
\begin{tabular}{l@{} c@{} c@{} c@{} c@{} c@{} c@{} c@{}}
\toprule
Retriever      & R@10   & \ R@100  & \ R@200  & \ R@400  & \ R@1k & \ R@2k & \ R@4k \\
\midrule
sparse\&dense & 0.61 & 0.81 & 0.86 & 0.89 & 0.93 & 0.94 & 0.95 \\
dense & 0.53 & 0.72 & 0.77 & 0.84 & 0.88 & 0.90 & 0.93 \\
sparse & 0.59 & 0.76 & 0.82 & 0.86 & 0.89 & 0.93 & 0.94 \\
\bottomrule
\end{tabular}
\caption{Recall of gold documents under single-document dataset using different retrieval methods.}
\label{tab:recall_single}
\end{table}

\textbf{Retrieval Setting:}
We utilize the provided \textit{Opensearch Sparse Index} as $M_{ret}^{sparse}$ and \textit{Pinecone Dense Index} as $M_{ret}^{dense}$, with intfloat/e5-base-v2 as the dense encoder, for retrieval. API requests are used to obtain results. Based on preliminary recall experiments and efficiency considerations shown in Table \ref{tab:recall_single} and \ref{tab:recall_multi}, we determined an optimal retrieval quantity of $N_{ret}=2000$.

\begin{table}[ht]
\centering
\footnotesize
\begin{tabular}{l@{} c@{} c@{} c@{} c@{} c@{} c@{} c@{} c@{}} 
\toprule
Model & Params & \ \ R@3 & \ \ R@5 & \ R@10  & \ R@20  & \ R@50  & \ R@400 \\
\midrule
None & - & 0.45 & 0.52 & 0.56 & 0.64 & 0.71 & 0.85 \\
bge-reranker-v2-m3 & 568M & 0.58 & 0.64 & 0.68 & 0.73 & 0.78 & 0.85 \\
jina-reranker-v2-base-multi & 278M & 0.53 & 0.61 & 0.69 & 0.74 & 0.82 & 0.91 \\
bge-reranker-v2-gemma & 2.51B & 0.69 & 0.76 & 0.81 & 0.85 & 0.90 & 0.93 \\
jina-reranker-m0 & 2.44B & 0.74 & 0.81 & 0.85 & 0.88 & 0.91 & 0.93 \\
jina-reranker-m0 (ft) & 2.44B & 0.76 & 0.80 & 0.83 & 0.86 & 0.91 & 0.94 \\
jina-reranker-m0 (ft-us) & 2.44B & 0.73 & 0.77 & 0.82 & 0.87 & 0.89 & 0.94 \\
\bottomrule
\end{tabular}
\caption{Comparison of ranking performance of different open-source ranking models on a subset of single-document dataset.}
\label{tab:reranker_single}
\end{table}

\begin{table}[ht]
\centering
\footnotesize
\begin{tabular}{l@{} c@{} c@{} c@{} c@{} c@{} c@{} c@{} c@{}} 
\toprule
Model & Params & \ \ R@3 & \ \ R@5 & \ R@10  & \ R@20  & \ R@50  & \ R@400 \\
\midrule
None & - & 0.24 & 0.32 & 0.39 & 0.47 & 0.57 & 0.73 \\
bge-reranker-v2-m3 & 568M & 0.26 & 0.33 & 0.41 & 0.49 & 0.60 & 0.77 \\
jina-reranker-v2-base-multi & 278M & 0.25 & 0.33 & 0.43 & 0.48 & 0.62 & 0.79 \\
bge-reranker-v2-gemma & 2.51B & 0.31 & 0.35 & 0.45 & 0.51 & 0.62 & 0.81 \\
jina-reranker-m0 & 2.44B & 0.36 & 0.43 & 0.53 & 0.62 & 0.72 & 0.87 \\
jina-reranker-m0 (ft) & 2.44B & 0.36 & 0.42 & 0.53 & 0.60 & 0.71 & 0.87 \\
jina-reranker-m0 (ft-us) & 2.44B & 0.39 & 0.45 & 0.54 & 0.62 & 0.72 & 0.87 \\
\bottomrule
\end{tabular}
\caption{Comparison of ranking performance of different open-source ranking models on a subset of  multi-document dataset. If only one gold document is recalled, the recall rate is 50\%.}
\label{tab:reranker_multi}
\end{table}

\textbf{Model Selection:} The LLM $M_{ans}$, $M_{summ}$ we use is the original Falcon3-10B-Instruct without further training, and the $M_{know}$ used is the Falcon3-10B-Instruct model fine-tuned via LoRA.
As it performed best in our preliminary reranking experiments, the reranking model $M_{rank}$ adopts the jina-reranker-m0 model, with detailed results shown in Table \ref{tab:reranker_single} and \ref{tab:reranker_multi}. We also fine-tuned the jina-reranker-m0 model using a training set generated by DataMorgana. We explored two positive-to-negative sample ratios: 1:16 (ft) and 4:16 (ft-us). However, the fine-tuned model's performance did not consistently surpass that of the original model. Further details are provided in the Appendix.

\begin{table}
\begin{tabular}{ccccc}
\toprule
$N_{ans}$ & \multicolumn{2}{c}{single-document} & \multicolumn{2}{c}{multi-document} \\ 
 & relevance & faithfulness & relevance & faithfulness \\ 
\midrule
30 & 1.23 & 0.36 & 1.17 & 0.24 \\
20 & 1.33 & 0.40 & 1.22 & 0.28 \\
10 & 1.42 & 0.48 & 1.27 & 0.30 \\
5 & 1.42 & 0.48 & 1.12 & 0.21 \\
3 & 1.40 & 0.48 & 1.06 & 0.16 \\ 
\bottomrule
\end{tabular}
\caption{The relationship between the number of reference documents $N_{ans}$ and the automatically evaluated score of answers when answering questions. We use Deepseek-v3 for scoring.}
\label{tab:answer_topk}
\end{table}

\textbf{$N_{ans}$ Selection:} Increasing the number of candidate answer documents $N_{ans}$ generally raises the probability of including the most relevant target document. However, a larger $N_{ans}$ also introduces more irrelevant documents, which can misguide the model and lead to sub-optimal answer generation. To evaluate this trade-off, we constructed evaluation prompts based on official metrics and used the DeepSeek-V3 \citep{liu2024deepseek} model to assess both relevance and faithfulness. The prompt is shown in Figure \ref{fig:prompt_eval} and the evaluation results are shown in Table \ref{tab:answer_topk}. Our comparative analysis across different $N_{ans}$ ans values demonstrates that $N_{ans}=10$ consistently achieves optimal performance for both question types examined.

\begin{table}[ht]
\centering
\small
\begin{tabular}{l c@{}c@{}c@{}c@{}c@{}c@{}}
\toprule
Method & $N_{know}$ & \ \ R@3 & \ \ R@5 & \ R@10 & \ R@20 \\
\midrule
jina-reranker-m0    & -      & 0.36    & 0.42    & 0.53    & 0.61    \\
Ours    & 0       & 0.34    & 0.40    & 0.52    & 0.60    \\
    & 2       & 0.35    & 0.42    & 0.53    & 0.63    \\
    & 5       & 0.36    & 0.44    & 0.55    & 0.64    \\
    & 8       & 0.35    & 0.43    & 0.54    & 0.63    \\
\bottomrule
\end{tabular}
\caption{Performance comparison between reranking of the proposed knowledge aware diverse reranking pipeline and using only the jina reranker in multi-document dataset.}
\label{tab:ours_main_multi}
\end{table}

\begin{table}[ht]
\centering
\small
\begin{tabular}{l c@{}c@{}c@{}c@{}c@{}c@{}}
\toprule
Method & $N_{know}$ & \ \ R@3 & \ \ R@5 & \ R@10 & \ R@20 \\
\midrule
jina-reranker-m0    & --      & 0.75    & 0.82    & 0.85    & 0.88    \\
Ours    & 0       & 0.71    & 0.76    & 0.82    & 0.85    \\
    & 2       & 0.74    & 0.82    & 0.84    & 0.88    \\
    & 5       & 0.75    & 0.83    & 0.85    & 0.87    \\
    & 8       & 0.73    & 0.79    & 0.84    & 0.87    \\
\bottomrule
\end{tabular}
\caption{Performance comparison between reranking of the proposed knowledge aware diverse reranking pipeline and using only the jina reranker in single-document dataset.}
\label{tab:ours_main_single}
\end{table}

\textbf{Parameters in Knowledge Element Declaration:} 
We test the performance of the knowledge aware diverse reranking method using varying numbers of $N_{know}$. As Table \ref{tab:ours_main_multi} and \ref{tab:ours_main_single} show, our approach with $N_{know}=5$ consistently outperforms the original Jina reranking model in multi-document scenarios, while showing no significant performance degradation in single-document settings. This demonstrates that diversity reranking effectively mitigates the \textbf{bias issue}. Notably, when the model generates new knowledge elements without referencing retrieved documents, performance generally falls below that of the Jina re-ranking model. This confirms that our proposed document-referencing mechanism alleviates the \textbf{misleading problem}.

\subsection{Inference Deployment}

To enhance the efficiency of the complete RAG workflow, we propose a cascaded producer-consumer framework. This design addresses the inherent sequential dependencies within our RAG process. The workflow is orchestrated into distinct, concurrently executing stages: retrieval, initial reranking, knowledge decomposition, knowledge summarization, further reranking, and answer generation. 
Each stage processes its input and queues the results for subsequent steps. We leverage Flask for deploying reranking model and vLLM for LLMs, ensuring a decoupled and scalable pipeline. 
Furthermore, for reranking and knowledge decomposition, we employ multi-process asynchronous task splitting during request processing, allowing multiple processes to concurrently handle partial tasks and integrate results in an ordered manner. This architecture enables dynamic resource allocation for API services based on performance bottlenecks, thereby optimizing overall efficiency.

\section{Results and Conclusion}

Our method achieved the highest scores in both Correctness and Faithfulness automatic evaluation metrics on the Session 2 dataset of the official online evaluation, as assessed by Claude-3.5 Sonnet (Table \ref{tab:combined_tables}). This superior performance also extends to the Session 1 dataset, where our approach outperformed other methods under the same question distribution.

This paper introduces Knowledge-Aware Diverse Reranking, a novel RAG framework designed to mitigate bias issue caused by reranking based on the original question and the misleading problem due to LLM's lack of understanding of the question context. The efficacy of our method is substantiated through comprehensive local experiments and online evaluations.

Due to temporal constraints, several design choices were not subjected to exhaustive empirical validation, and numerous initial, less successful, attempts were not further optimized. These aspects present opportunities for future investigations within the broader RAG domain and for subsequent iterations of the LiveRAG competition. A more detailed discussion of these unexplored avenues and unrefined explorations is provided in the Appendix.

\bibliographystyle{ACM-Reference-Format}
\bibliography{sample-base}


\begin{thebibliography}{11}


\ifx \showCODEN    \undefined \def \showCODEN     #1{\unskip}     \fi
\ifx \showISBNx    \undefined \def \showISBNx     #1{\unskip}     \fi
\ifx \showISBNxiii \undefined \def \showISBNxiii  #1{\unskip}     \fi
\ifx \showISSN     \undefined \def \showISSN      #1{\unskip}     \fi
\ifx \showLCCN     \undefined \def \showLCCN      #1{\unskip}     \fi
\ifx \shownote     \undefined \def \shownote      #1{#1}          \fi
\ifx \showarticletitle \undefined \def \showarticletitle #1{#1}   \fi
\ifx \showURL      \undefined \def \showURL       {\relax}        \fi
\providecommand\bibfield[2]{#2}
\providecommand\bibinfo[2]{#2}
\providecommand\natexlab[1]{#1}
\providecommand\showeprint[2][]{arXiv:#2}

\bibitem[Fan et~al\mbox{.}(2024)]%
        {fan2024survey}
\bibfield{author}{\bibinfo{person}{Wenqi Fan}, \bibinfo{person}{Yujuan Ding}, \bibinfo{person}{Liangbo Ning}, \bibinfo{person}{Shijie Wang}, \bibinfo{person}{Hengyun Li}, \bibinfo{person}{Dawei Yin}, \bibinfo{person}{Tat-Seng Chua}, {and} \bibinfo{person}{Qing Li}.} \bibinfo{year}{2024}\natexlab{}.
\newblock \showarticletitle{A survey on rag meeting llms: Towards retrieval-augmented large language models}. In \bibinfo{booktitle}{\emph{Proceedings of the 30th ACM SIGKDD Conference on Knowledge Discovery and Data Mining}}. \bibinfo{pages}{6491--6501}.
\newblock


\bibitem[Filice et~al\mbox{.}(2025)]%
        {filice2025generating}
\bibfield{author}{\bibinfo{person}{Simone Filice}, \bibinfo{person}{Guy Horowitz}, \bibinfo{person}{David Carmel}, \bibinfo{person}{Zohar Karnin}, \bibinfo{person}{Liane Lewin-Eytan}, {and} \bibinfo{person}{Yoelle Maarek}.} \bibinfo{year}{2025}\natexlab{}.
\newblock \showarticletitle{Generating Diverse Q\&A Benchmarks for RAG Evaluation with DataMorgana}.
\newblock \bibinfo{journal}{\emph{arXiv preprint arXiv:2501.12789}} (\bibinfo{year}{2025}).
\newblock


\bibitem[Gao et~al\mbox{.}(2023)]%
        {gao2023retrieval}
\bibfield{author}{\bibinfo{person}{Yunfan Gao}, \bibinfo{person}{Yun Xiong}, \bibinfo{person}{Xinyu Gao}, \bibinfo{person}{Kangxiang Jia}, \bibinfo{person}{Jinliu Pan}, \bibinfo{person}{Yuxi Bi}, \bibinfo{person}{Yixin Dai}, \bibinfo{person}{Jiawei Sun}, \bibinfo{person}{Haofen Wang}, {and} \bibinfo{person}{Haofen Wang}.} \bibinfo{year}{2023}\natexlab{}.
\newblock \showarticletitle{Retrieval-augmented generation for large language models: A survey}.
\newblock \bibinfo{journal}{\emph{arXiv preprint arXiv:2312.10997}} \bibinfo{volume}{2}, \bibinfo{number}{1} (\bibinfo{year}{2023}).
\newblock


\bibitem[Guo et~al\mbox{.}(2025)]%
        {guo2025deepseek}
\bibfield{author}{\bibinfo{person}{Daya Guo}, \bibinfo{person}{Dejian Yang}, \bibinfo{person}{Haowei Zhang}, \bibinfo{person}{Junxiao Song}, \bibinfo{person}{Ruoyu Zhang}, \bibinfo{person}{Runxin Xu}, \bibinfo{person}{Qihao Zhu}, \bibinfo{person}{Shirong Ma}, \bibinfo{person}{Peiyi Wang}, \bibinfo{person}{Xiao Bi}, {et~al\mbox{.}}} \bibinfo{year}{2025}\natexlab{}.
\newblock \showarticletitle{Deepseek-r1: Incentivizing reasoning capability in llms via reinforcement learning}.
\newblock \bibinfo{journal}{\emph{arXiv preprint arXiv:2501.12948}} (\bibinfo{year}{2025}).
\newblock


\bibitem[Hu et~al\mbox{.}(2022)]%
        {hu2022lora}
\bibfield{author}{\bibinfo{person}{Edward~J Hu}, \bibinfo{person}{Yelong Shen}, \bibinfo{person}{Phillip Wallis}, \bibinfo{person}{Zeyuan Allen-Zhu}, \bibinfo{person}{Yuanzhi Li}, \bibinfo{person}{Shean Wang}, \bibinfo{person}{Lu Wang}, \bibinfo{person}{Weizhu Chen}, {et~al\mbox{.}}} \bibinfo{year}{2022}\natexlab{}.
\newblock \showarticletitle{Lora: Low-rank adaptation of large language models.}
\newblock \bibinfo{journal}{\emph{ICLR}} \bibinfo{volume}{1}, \bibinfo{number}{2} (\bibinfo{year}{2022}), \bibinfo{pages}{3}.
\newblock


\bibitem[Khaki et~al\mbox{.}(2024)]%
        {khaki2024rs}
\bibfield{author}{\bibinfo{person}{Saeed Khaki}, \bibinfo{person}{JinJin Li}, \bibinfo{person}{Lan Ma}, \bibinfo{person}{Liu Yang}, {and} \bibinfo{person}{Prathap Ramachandra}.} \bibinfo{year}{2024}\natexlab{}.
\newblock \showarticletitle{Rs-dpo: A hybrid rejection sampling and direct preference optimization method for alignment of large language models}.
\newblock \bibinfo{journal}{\emph{arXiv preprint arXiv:2402.10038}} (\bibinfo{year}{2024}).
\newblock


\bibitem[Liu et~al\mbox{.}(2024)]%
        {liu2024deepseek}
\bibfield{author}{\bibinfo{person}{Aixin Liu}, \bibinfo{person}{Bei Feng}, \bibinfo{person}{Bing Xue}, \bibinfo{person}{Bingxuan Wang}, \bibinfo{person}{Bochao Wu}, \bibinfo{person}{Chengda Lu}, \bibinfo{person}{Chenggang Zhao}, \bibinfo{person}{Chengqi Deng}, \bibinfo{person}{Chenyu Zhang}, \bibinfo{person}{Chong Ruan}, {et~al\mbox{.}}} \bibinfo{year}{2024}\natexlab{}.
\newblock \showarticletitle{Deepseek-v3 technical report}.
\newblock \bibinfo{journal}{\emph{arXiv preprint arXiv:2412.19437}} (\bibinfo{year}{2024}).
\newblock


\bibitem[Penedo et~al\mbox{.}(2024)]%
        {penedo2024the}
\bibfield{author}{\bibinfo{person}{Guilherme Penedo}, \bibinfo{person}{Hynek Kydl{\'\i}{\v{c}}ek}, \bibinfo{person}{Loubna~Ben allal}, \bibinfo{person}{Anton Lozhkov}, \bibinfo{person}{Margaret Mitchell}, \bibinfo{person}{Colin Raffel}, \bibinfo{person}{Leandro~Von Werra}, {and} \bibinfo{person}{Thomas Wolf}.} \bibinfo{year}{2024}\natexlab{}.
\newblock \showarticletitle{The FineWeb Datasets: Decanting the Web for the Finest Text Data at Scale}. In \bibinfo{booktitle}{\emph{The Thirty-eight Conference on Neural Information Processing Systems Datasets and Benchmarks Track}}.
\newblock
\urldef\tempurl%
\url{https://openreview.net/forum?id=n6SCkn2QaG}
\showURL{%
\tempurl}


\bibitem[Team(2024)]%
        {Falcon3}
\bibfield{author}{\bibinfo{person}{TII Team}.} \bibinfo{year}{2024}\natexlab{}.
\newblock \bibinfo{title}{The Falcon 3 family of Open Models}.
\newblock


\bibitem[Tong et~al\mbox{.}(2024)]%
        {tong2024dart}
\bibfield{author}{\bibinfo{person}{Yuxuan Tong}, \bibinfo{person}{Xiwen Zhang}, \bibinfo{person}{Rui Wang}, \bibinfo{person}{Ruidong Wu}, {and} \bibinfo{person}{Junxian He}.} \bibinfo{year}{2024}\natexlab{}.
\newblock \showarticletitle{Dart-math: Difficulty-aware rejection tuning for mathematical problem-solving}.
\newblock \bibinfo{journal}{\emph{Advances in Neural Information Processing Systems}}  \bibinfo{volume}{37} (\bibinfo{year}{2024}), \bibinfo{pages}{7821--7846}.
\newblock


\bibitem[Wang et~al\mbox{.}(2022)]%
        {wang2022text}
\bibfield{author}{\bibinfo{person}{Liang Wang}, \bibinfo{person}{Nan Yang}, \bibinfo{person}{Xiaolong Huang}, \bibinfo{person}{Binxing Jiao}, \bibinfo{person}{Linjun Yang}, \bibinfo{person}{Daxin Jiang}, \bibinfo{person}{Rangan Majumder}, {and} \bibinfo{person}{Furu Wei}.} \bibinfo{year}{2022}\natexlab{}.
\newblock \showarticletitle{Text Embeddings by Weakly-Supervised Contrastive Pre-training}.
\newblock \bibinfo{journal}{\emph{arXiv preprint arXiv:2212.03533}} (\bibinfo{year}{2022}).
\newblock


\end{thebibliography}


\newpage

\appendix
\section{Leaderboard}

See Table \ref{tab:combined_tables}
\begin{table*}[ht]
    \centering
    \begin{subtable}[t]{0.48\textwidth}
        \centering
        \small
        \begin{tabular}{lrr}
            \toprule
            \textbf{Team Name} & \textbf{Correctness} & \textbf{Faithfulness} \\
            \midrule
            RMIT-ADMS & 1.199317 & 0.477382 \\
            RUC\_DeepSearch & 0.969273 & 0.387808 \\
            Ped100X & 0.928893 & 0.043381 \\
            PRMAS-DRCA & 0.922780 & 0.410600 \\
            Hybrid Search with Graph & 0.875091 & 0.315802 \\
            BagBag & 0.694073 & -0.911353 \\
            UniClustRAG & 0.685146 & 0.460062 \\
            METURAG & 0.673451 & 0.325339 \\
            DeepRAG & 0.566053 & 0.097828 \\
            UIS-IAI & 0.552328 & 0.433697 \\
            SNU-LDILab & 0.517367 & 0.103027 \\
            Gravitational Lens & 0.376637 & -0.988097 \\
            - & - & - \\
            \bottomrule
        \end{tabular}
        \caption{Session 1 Results}
        
    \end{subtable}
    \hfill
    \begin{subtable}[t]{0.48\textwidth}
        \centering
        \small
        \begin{tabular}{lrr}
            \toprule
            \textbf{Team Name} & \textbf{Correctness} & \textbf{Faithfulness} \\
            \midrule
            Magikarp (Ours) & \textbf{1.231578} & \textbf{0.656464} \\
            UDInfo & 1.200586 & 0.623175 \\
            RAGtifier & 1.134454 & 0.552365 \\
            HLTCOE & 1.070111 & 0.340711 \\
            Ragmatazz & 1.011956 & 0.519394 \\
            ScaledRAG & 0.996348 & 0.418273 \\
            Emorag & 0.890718 & 0.556581 \\
            Graph-Enhanced RAG & 0.875714 & 0.529335 \\
            Multi-Agent Adaptive RAG & 0.836110 & 0.200420 \\
            Starlight & 0.818337 & 0.433003 \\
            NoobRAG & 0.655292 & 0.154648 \\
            UIUC-RAGents & 0.565043 & -0.302616 \\
            AugmentRAG-TUD & 0.532533 & 0.655634 \\
            \bottomrule
        \end{tabular}
        \caption{Session 2 Results}
        
    \end{subtable}
    \caption{Comparison of Different RAG Systems with automatic evaluation. Teams ranked by the correctness of the answers.}
    \label{tab:combined_tables}
\end{table*}

\section{Exploring Unsuccessful Attempts}

This section details various unsuccessful strategies we explored to improve retrieval performance, focusing on query rewriting and reranking techniques. These attempts, while not yielding significant improvements in our specific settings, offer valuable insights and directions for future research.

\subsection{Query Rewriting with Falcon3-10B-Instruct}

To enhance retrieval recall, we experimented with endowing the Falcon3-10B-Instruct model with query rewriting capabilities through supervised fine-tuning. We explored several approaches for constructing training data:
(1) We leveraged DeepSeek-v3 to extract the most valuable keywords from questions, standard answers, and gold documents. The model was then trained to generate these keywords given a question.
(2) To alleviate the capability gap in the distillation process, we used Falcon3-10B-Instruct to generate a thought process and valuable retrieval keywords based on the input question. We sampled 16 iterations and selected the thought process and keywords that yielded the best retrieval performance as training data.

Beyond keywords, we also attempted to generate pseudo-documents or pseudo-answer. However, none of these query rewriting methods outperformed using the original query with a sparse-dense hybrid retrieval approach. 
We hypothesize that for single question retrieval scenarios, query rewriting heavily relies on the model's parametric knowledge. 
Moreover, rewriting might inadvertently steer the query towards a specific, potentially suboptimal, search direction, leading to performance degradation. It's also possible that the SFT approach limited the model's full potential, and reinforcement learning might offer a more effective training paradigm.

\subsection{Fine-tuning Jina Reranker}

We attempted to fine-tune jina-reranker-m0 model using data generated by DM. We meticulously designed 16 types of hard negative samples, including those with high ROUGE-L scores relative to the answer, question, and gold document, as well as the most relevant documents from dense and sparse retrieval. 
Training was conducted using contrastive loss or BCE loss, but no significant performance improvement was observed. A potential reason for this outcome is our labeled dataset misclassification of potentially relevant documents as negative samples, which could confuse the model during training. Simpler negative sampling strategies or larger-scale training datasets might be promising avenues for future exploration.

\subsection{LLM-based Reranking}

We explored a Chain-of-Thought approach using the Falcon3-10B-Instruct model to assess the value of each document on top of an initial reranking. Documents deemed "False" by the LLM had their scores reduced by 1. We distilled DeepSeek-v3's CoT data for relevance judgment to train this mechanism.

Experiments indicated that this method improved R@3 and R@5 metrics in single-document scenarios but showed no benefit for multi-document questions. Furthermore, this approach introduced considerable inference latency, precluding its adoption in the final solution. 
In contrast to query rewriting, which demands additional knowledge, reranking solely relies on matching and judging based on provided information. This makes it a more promising optimization direction for RAG systems built upon LLMs with parametric limitations. 
Future work should explore more generalized reinforcement learning training methods for reranking. Additionally, the presence of false positives and false negatives could lead to ranking instability; therefore, future research should also focus on soft-combining the judgment results of LLMs with reranker scores.

\subsection{Knowledge Gap-based Reranking}

Our initial motivation for designing a knowledge decomposition strategy was to identify and reorder documents based on knowledge gaps within the top-ranked results. Specifically, we used a knowledge declaration model to identify missing knowledge elements from each of the top 5 documents (i.e., total output knowledge minus existing knowledge in the current document). These missing knowledge elements were then used to rerank documents outside the initial top 5.

However, even in multi-document scenarios, this method did not outperform using only Jina's reranking model in our experiments. We believe the concept of a "knowledge gap" aligns with advanced methods like DeepSearch. Nevertheless, LLMs struggle to accurately quantify "existing knowledge within the current document," especially when dealing with open-ended answers, as this standard is inherently vague and challenging to define. Future improvements should focus on employing more robust reasoning models capable of performing simultaneous reasoning and searching.

\section{Prompt and examples}

\begin{figure}
\begin{promptbox}
\textbf{System:}\\
Please first analyze the given problem and determine which knowledge elements are required to answer the given problem.\\
Please first think step by step and then output in numbered list (no more than 4 points).\\
Then analyze which knowledge points are provided in the given document from these required knowledge elements.\\
Please first think step by step and then output the number of knowlege elements (if no knowledge elements are provided, output None).\\

**Example output format:**\\
Thoughts knowlege requirements: \\
<your thoughts here>\\

Knowledge Elements: \\
<numbered list>\\

Analysis for given document: \\
<compare and analyse document and above knowledge point by point>\\

Given Knowledge: \\
<selected numbers> \\

Example Input:\\
\{demo\_input\}\\
Example Output:\\
\{demo\_output\}\\

\textbf{User:}\\
Question: \{question\}\\
Document: \{document\}
\end{promptbox}
\caption{The prompt for Knowledge Element Declaration.}
\label{fig:prompt_dec}
\end{figure}

\begin{figure}
\begin{promptbox}
\textbf{System:}\\
You have identified the knowledge elements required to answer the given question based on the provided question and the retrieved relevant documents.\\
Please select two of the most important, complete, and non-redundant knowledge elements from the identified knowledge elements for further retrieving the knowledge base to answer the given question.\\
Please think step by step first, and finally output the result in the format of a Python list.\\

**Example Output Format:**\\
Thoughts: \\
<your thoughts here>\\

Selected Knowledge Elements:
\begin{verbatim}
```json
[
    "Knowledge Element 1",
    "Knowledge Element 2"
]
```
\end{verbatim}

\#\# Examples:\\
**Example Input:**\\
\{demo\_input\}\\

**Example Output:**\\
\{demo\_output\}\\

\textbf{User:}\\

Question: \{question\}\\
Knowledge Elements:
\begin{verbatim}
```
{knowledge_elements}
```
\end{verbatim}
\end{promptbox}
\caption{The prompt for knowledge element summarization.}
\label{fig:prompt_summ}

\end{figure}




\begin{figure}
\begin{promptbox}
This is the question type used for generating RAG (Retrieval-Augmented Generation) questions. The entire list captures various dimensions of questions, including factuality, premise, phrasing, and linguistic\_variation. Each dimensional aspect further contains parallel options and possibilities; for example, regarding factuality, there are two types: factoid and open-ended. Please continue brainstorming new categorization names and their corresponding categories.\\

\{question\_categorizations\}

\end{promptbox}
\caption{The prompt for new question type brainstorming.}
\label{fig:prompt_brain}
\end{figure}

\begin{figure*}
\begin{widepromptbox}
\textbf{Input:} \\

Question: What are the key differences between getting lost in the backcountry and being stranded without man-made resources in terms of the survival skills required to handle each situation?\\

Document: Bushcraft’ is a word that gets thrown around very often in the survival community, but it’s also a word that far fewer people understand it. A truly skilled survivalist is someone who can use resources provided by nature exclusively to survive. Ask yourself this: if you were stranded out in the wilderness tomorrow with nothing but the clothes on your back and could only use completely natural resources to survive, would you be able to? If your honest answer is no, then you will probably find the information presented in this article useful. We are going to provide you with a definitive list of bushcraft skills that will allow you to survive in the wilderness using no man-made materials whatsoever. THE BOW DRILL METHOD Everyone knows how important fire is in any survival situation. But not everyone is capable of starting a fire without a flint striker, lighter, or matches. It’s imperative that you learn a way to start a fire without any of those kinds of fire starting devices. The best method for starting a fire without any man-made materials is the bow drill method. This method requires you to collect a flat piece of wood (to serve as the fireboard) with a notch cut into it. You also need a bow complete with a vine for the cord, and a sharpened stick as the spindle. Wrap the vine around the spindle and place the point of the spindle right above where the notch is in the fireboard. Proceed to run the bow back and forth very quickly and over an extended period, friction and heat build between the spindle and board. The small shavings of wood will then fall into the notch. Have a tinder nest already made and different sized kindling on standby. Once you get an ember or smoke, you can transfer the ember into the tinder next to get your fire started. Proceed to add kindling, and you’ll have a fire going. The bow drill method may sound simple enough on paper, but it’s going to be more physically taxing in real life. You never want a true survival situation to be the first time you practice the bow drill method of fire starting. For this reason, practice extensively now on weekends or whenever you have the time until you become a master at it. That way, it will seem virtually second nature to you in a life-or-death situation. The skill of tying together two strips of vine or other man-made materials is one that you will not only use in a survival situation but throughout your life as well\\

\textbf{Output:}\\

Thoughts knowledge requirements: \\
To answer the question about the key differences between getting lost in the backcountry and being stranded without man-made resources, we need to understand the specific survival skills required for each scenario. Getting lost in the backcountry might involve using some man-made resources (e.g., a map, compass, or gear), while being stranded without man-made resources requires pure bushcraft skills (e.g., fire-starting, shelter-building, foraging). The document focuses on bushcraft skills, which are more relevant to the latter scenario.\\

Knowledge Elements:\\
1. Understanding the definition and scope of bushcraft (survival using only natural resources).\\
2. Fire-starting techniques without man-made tools (e.g., bow drill method).\\
3. The importance of practicing bushcraft skills before a survival situation.\\
4. General survival skills like tying materials (though not explicitly detailed for shelter or foraging).\\

Analysis for given document: \\
The document primarily discusses bushcraft skills, emphasizing fire-starting without man-made tools (bow drill method) and the need for practice. It does not explicitly contrast backcountry survival (with potential man-made resources) vs. stranded survival (without man-made resources), but it provides insights into the latter. The document covers points 1, 2, and 3 but does not delve into broader survival skill comparisons (e.g., navigation, shelter-building for backcountry vs. bushcraft).\\

Given Knowledge: \\
1, 2, 3\\
\end{widepromptbox}
\caption{A Demonstration of Knowledge Element Declaration.}
\label{fig:demo_dec}
\end{figure*}

\begin{figure*}
\begin{widepromptbox}
\textbf{Input:} \\
Question: How has the understanding of decision-making and choice abundance evolved from Barry Schwartz's early research on the jam experiment to more recent psychological approaches like the U-Theory?\\

Knowledge Elements: 
\begin{verbatim}
```
Barry Schwartz's jam experiment and its findings on choice and decision-making.
The adverse effects of choice abundance, including analysis paralysis, buyer's remorse, and decision fatigue.
The U-Theory and its relevance to understanding decision-making in the context of choice abundance.
The impact of choice abundance on consumer behavior, conversions, retention, and revenue.
Barry Schwartz's jam experiment and its implications for decision-making.
The concept of choice overload and its effects on decision-making.
The role of cognitive limits in decision-making (e.g., George Miller's "magical number seven").
The impact of information quantity on purchasing decisions (Iyengar and Lepper's research).
Barry Schwartz's jam experiment and its impact on understanding choice overload.
The concept of choice abundance and its psychological effects.
The evolution of psychological theories on decision-making.
Introduction to U-Theory and its application in understanding decision-making.
Barry Schwartz's jam experiment and its findings on decision-making paralysis.
The concept of choice abundance and its effects on consumer behavior.
U-Theory's approach to decision-making, emphasizing the balance between too few and too many options.
Evolution of psychological theories on decision-making from Schwartz's early work to U-Theory.
Barry Schwartz's jam experiment and its findings on decision-making paralysis.
The concept of choice abundance and its effects on consumer behavior.
U-Theory's perspective on decision-making and its differences from Schwartz's early work.
Evolution of psychological approaches to decision-making and choice abundance.
```
\end{verbatim}

\textbf{Output:}
Thoughts:\\
The question asks about the evolution from Barry Schwartz's jam experiment to U-Theory. The most critical points are the foundational findings of the jam experiment (highlighting choice overload) and U-Theory's modern approach to balancing choice abundance. These directly address the "evolution" in understanding, avoiding redundancy with other supporting concepts like cognitive limits or consumer behavior impacts.\\

Selected Knowledge Elements: 
\begin{verbatim}
```json
[
    "Barry Schwartz's jam experiment and its findings on decision-making paralysis.",
    "U-Theory's approach to decision-making, emphasizing the balance between too few and too many options."
]
```
\end{verbatim}

\end{widepromptbox}
\caption{A Demonstration of Knowledge Element Summarization.}
\label{fig:demo_summ}

\end{figure*}

\begin{figure*}
\begin{widepromptbox}
\textbf{System:} \\
You are an QA evaluation assistant.\\
Your task is assess the quality of the answer provided by the model base on metrics **Relevance** and **Faithfulness**.\\
You are given a question, one or two reference documents, the ground truth answer and the model's output.\\

Please first evaluate the relevance of the output with respect to the question and the ground truth answer.
And then evaluate the faithfulness of the output with respect to the documents.\\

The definition of **Relevance** score is as follows:\\
Combines elements of equivalence (semantic match with ground truth) and relevance (degree to which the answer directly addresses the question).\\
Graded on a four-point scale:\\
2: Correct and relevant (no irrelevant information).\\
1: Correct but contains irrelevant information.\\
0: No answer provided (abstention).\\
-1: Incorrect answer.\\

The definition of **Faithfulness** score is as follows:\\
Assesses whether the response is grounded in the retrieved passages.\\
Graded on a three-point scale:\\
1: Full support. All answer parts are grounded.\\
0: Partial support. Not all answer parts are grounded.\\
-1: No support. All answer parts are not grounded.\\

Please first think step by step and then output your evaluation scores in a json format.\\
Example output format:\\
<your thoughts here>\\
\begin{verbatim}
```json
{
    "relevance": <your assessment of relevance>,
    "faithfulness": <your assessment of faithfulness>
}
```
\end{verbatim}

\textbf{User:}\\
**Question:** \{question\}\\
**Ground Truth Answer:** \{answer\}\\
**Reference Documents:** \{gold\_context\}\\
**Model's Output:** \{output\}\\

\end{widepromptbox}
\caption{Our prompt for answer scoring using DeepSeek-v3.}
\label{fig:prompt_eval}

\end{figure*}

\begin{table*}[ht]
\centering

\begin{tabular}{llp{7.5cm}r}
\toprule
\textbf{Categorization} & \textbf{Category} & \textbf{Description} & \textbf{Probability} \\
\midrule
\multirow{2}{*}{Factuality} 
    & factoid & Question seeking a specific, concise piece of information or a short fact about a particular subject, such as a name, date, or number. & 0.5 \\
    & open-ended & Question inviting detailed or exploratory responses, encouraging discussion or elaboration. & 0.5 \\
\addlinespace

\multirow{2}{*}{Premise} 
    & direct & Question that does not contain any premise or any information about the user. & 0.95 \\
    & with-premise & Question starting with a very short premise, where the user reveals their needs or some information about himself. & 0.05 \\
\addlinespace

\multirow{4}{*}{Phrasing} 
    & concise-and-natural & Phrased in the way people typically speak, reflecting everyday language use, without formal or artificial structure. It is a concise direct question consisting of less than 10 words. & 0.25 \\
    & verbose-and-natural & Phrased in the way people typically speak, reflecting everyday language use, without formal or artificial structure. It is a relatively long question consisting of more than 10 words and less than 20 words. & 0.75 \\
    & short-search-query & Phrased as a typed web query for search engines (only keywords, without punctuation and without a natural-sounding structure). It consists of less than 7 words. & 0.05 \\
    & long-search-query & Phrased as a typed web query for search engines (only keywords, without punctuation and without a natural-sounding structure). It consists of more than 6 words. & 0.05 \\
\addlinespace

\multirow{2}{*}{Linguistic Variation} 
    & similar-to-document & Phrased using the same terminology and phrases appearing in the document. & 0.5 \\
    & distant-from-document & Phrased using terms completely different from the ones appearing in the document. & 0.5 \\
\addlinespace

\multirow{3}{*}{Answer Format} 
    & numeric & Question expecting a numerical answer (e.g., dates, quantities, percentages). & 0.2 \\
    & textual & Question expecting a natural-language sentence or paragraph as an answer. & 0.5 \\
    & list & Question expecting a structured list of items (e.g., steps, features, categories). & 0.3 \\
\addlinespace

\multirow{2}{*}{Question Complexity} 
    & simple\_lookup & Requires finding a single, explicitly stated piece of information within the document. & 0.5 \\
    & synthesis & Requires combining information from multiple sentences or sections of the document. & 0.5 \\
\addlinespace

\multirow{2}{*}{Scope} 
    & narrow & Focuses on a very specific detail, entity, attribute, or event. & 0.5 \\
    & broad & Asks about a general theme, summary, main idea, or category. & 0.5 \\
\addlinespace

\multirow{4}{*}{Temporal Focus} 
    & specific-time-point & Refers to a single date, year, or specific event time. & 0.2 \\
    & time-duration & Refers to a span or period of time. & 0.15 \\
    & relative-time & Refers to time relative to another event or the present. & 0.15 \\
    & atemporal & No specific time focus, asking about general facts or processes. & 0.5 \\
\bottomrule
\end{tabular}
\caption{Question classification taxonomy for single-document.}
\label{tab:question_single}
\end{table*}

\begin{table*}[ht]
\centering
\begin{tabular}{llp{7.5cm}r}
\toprule
\textbf{Categorization} & \textbf{Category} & \textbf{Description} & \textbf{Probability} \\
\midrule
\multirow{7}{*}{Multi-doc}
    & comparison & A comparison question that requires comparing two related concepts or entities. The comparison must be natural and reasonable, i.e., comparing two entities by a common attribute which is meaningful and relevant to both entities. Example: ``Who is older, Glenn Hughes or Ross Lynch?'' Requires information from two documents about respective entities. & 0.3 \\
    
    & multi-aspect & A question about two different aspects of the same entity/concept. Example: ``What are the advantages of AI-powered diagnostics, and what are the associated risks?'' Requires two documents about different aspects. & 0.3 \\
    
    & cause-effect & Analysis of causal relationships where cause and effect are in separate documents. Example: ``How did the 2023 El Niño influence global agricultural yields?'' First document details causes, second describes effects. & 0.05 \\
    
    & conflicting-views & Addresses divergent opinions from different sources. Example: ``How do IPCC and Heartland Institute differ in sea level rise projections?'' Requires two documents with opposing perspectives. & 0.05 \\
    
    & definition-combination & Requires synthesizing two complementary definitions. Example: ``Combined definition of 'greenwashing' from policy documents and corporate reports.'' Merges partial definitions from two sources. & 0.1 \\
    
    & temporal-change & Asks about changes between time periods. Example: ``Gray wolf population in Yellowstone before (1995) and after (2015) reintroduction.'' Requires documents from different time periods. & 0.1 \\
    
    & multi-hop & Requires multi-step reasoning across documents. Example: ``What was the first company founded by Donald Trump's father?'' Needs sequential information retrieval. & 0.1 \\
\bottomrule
\end{tabular}
\caption{Question classification taxonomy for multi-document.}
\label{tab:question_multi}
\end{table*}

\end{document}